\documentclass[letterpaper]{article} 

\ifdefined\aaaianonymous
    \usepackage[submission]{aaai2026}  
\else
    \usepackage{aaai2026}              
\fi
\usepackage{comment}
\usepackage{times}  
\usepackage{helvet}  
\usepackage{courier}  
\usepackage[hyphens]{url}  
\usepackage{graphicx} 
\usepackage{natbib}  
\usepackage{caption} 
\usepackage{algorithm}
\usepackage{algorithmic}

\usepackage{newfloat}
\usepackage{listings}
\DeclareCaptionStyle{ruled}{labelfont=normalfont,labelsep=colon,strut=off} 
\floatstyle{ruled}
\newfloat{listing}{tb}{lst}{}
\floatname{listing}{Listing}

\usepackage{booktabs} 
\usepackage{siunitx} 
\usepackage{multirow}
\usepackage{pifont}
\usepackage{amssymb}
\usepackage{amsmath}

\usepackage{xcolor}

\ifdefined\aaaianonymous
    \title{MambaSeg: Harnessing Mamba for Accurate and Efficient Image-Event Semantic Segmentation}
\else
    \title{MambaSeg: Harnessing Mamba for Accurate and Efficient Image-Event Semantic Segmentation}
\fi

\author{
    Fuqiang Gu\textsuperscript{\rm 1,2}, Yuanke Li\textsuperscript{\rm 2},   Xianlei Long\textsuperscript{\rm 1}\thanks{Corresponding author: \textit{Xianlei Long and Zhenliang Ni.}}, Kangping Ji\textsuperscript{\rm 2}, Chao~Chen\textsuperscript{\rm 1}, Qingyi~Gu\textsuperscript{\rm 3}, Zhenliang Ni\textsuperscript{\rm 3}\footnotemark[1]
}
\affiliations{
    \textsuperscript{\rm 1}College of Computer Science, Chongqing University\\
    \textsuperscript{\rm 2}National Elite Institute of Engineering, Chongqing University\\
    \textsuperscript{\rm 3}Institute of Automation, Chinese Academy of Sciences\\
    gufq@cqu.edu.cn, lyk@cqu.edu.cn, xianlei.long@cqu.edu.cn,  jkp@stu.cqu.edu.cn, cschaochen@cqu.edu.cn, qingyi.gu@ia.ac.cn, nizhenliang@outlook.com
}

\usepackage{bibentry}

\begin{document}

\maketitle

\begin{abstract}
Semantic segmentation is a fundamental task in computer vision with wide-ranging applications, including autonomous driving and robotics. While RGB-based methods have achieved strong performance with CNNs and Transformers, their effectiveness degrades under fast motion, low-light, or high dynamic range conditions due to limitations of frame cameras. Event cameras offer complementary advantages such as high temporal resolution and low latency, yet lack color and texture, making them insufficient on their own. To address this, recent research has explored multimodal fusion of RGB and event data; however, many existing approaches are computationally expensive and focus primarily on spatial fusion, neglecting the temporal dynamics inherent in event streams.
In this work, we propose MambaSeg, a novel dual-branch semantic segmentation framework that employs parallel Mamba encoders to efficiently model RGB images and event streams. To reduce cross-modal ambiguity, we introduce the Dual-Dimensional Interaction Module (DDIM), comprising a Cross-Spatial Interaction Module (CSIM) and a Cross-Temporal Interaction Module (CTIM), which jointly perform fine-grained fusion along both spatial and temporal dimensions. This design improves cross-modal alignment, reduces ambiguity, and leverages the complementary properties of each modality. Extensive experiments on the DDD17 and DSEC datasets demonstrate that MambaSeg achieves state-of-the-art segmentation performance while significantly reducing computational cost, showcasing its promise for efficient, scalable, and robust multimodal perception.
\end{abstract}

\ifdefined\aaaianonymous
\else
\begin{links}
    \link{Code}{https://github.com/CQU-UISC/MambaSeg}
\end{links}
\fi

\section{Introduction}
Semantic segmentation is a fundamental task in computer vision with broad applications in autonomous driving, robotics, and scene understanding~\cite{lateef2019survey}. Driven by advances in convolutional neural networks (CNNs) and Transformers, RGB image-based methods have achieved significant progress~\cite{badrinarayanan2017segnet,xie2021segformer,ni2024context,ma2024ssa}. However, their reliance on conventional image sensors poses inherent limitations under adverse conditions such as high-speed motion, low illumination, and high dynamic range scenes, where motion blur and latency severely degrade performance~\cite{rebecq2019high}. 

Event cameras offer a compelling alternative by asynchronously capturing per-pixel intensity changes with microsecond latency, high temporal resolution, and wide dynamic range~\cite{gallego2020event}. These properties make them particularly well-suited for dynamic and low-light environments~\cite{gehrig2024low}. Yet, event streams lack color and fine-grained texture information, limiting their standalone utility in dense prediction tasks such as semantic segmentation~\cite{sun2022ess,jia2023event}. To harness the complementary strengths of both modalities, recent research has focused on RGB-event fusion techniques that combine the spatial richness of images with the temporal precision of events.

\begin{figure}[!t]
\centering
\includegraphics[width=0.85\columnwidth]{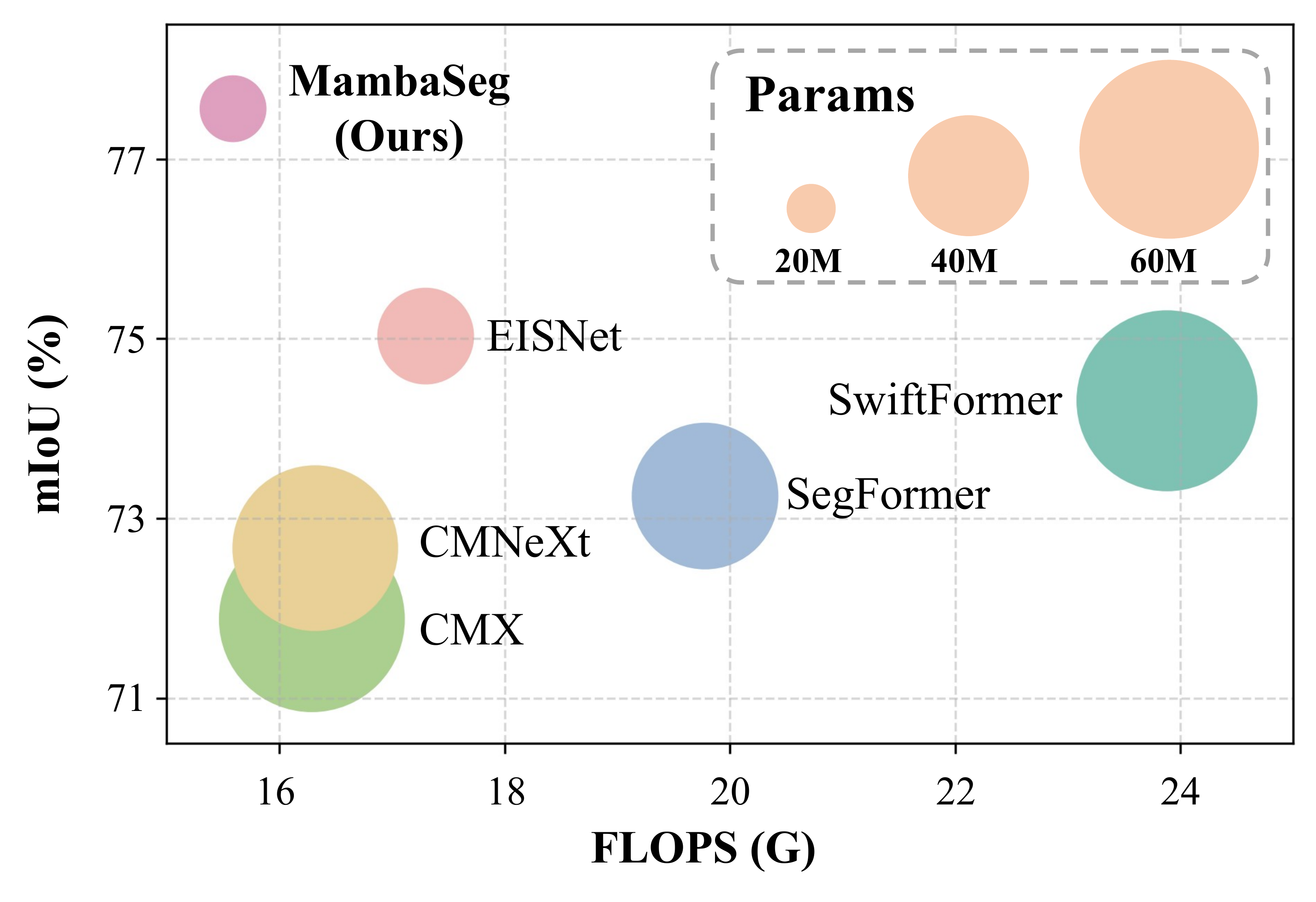}
\caption{Comparison of event-image fusion segmentation methods in terms of mIoU (\%), Parameters (M), and  multiply-accumulate operations (G).}
\label{fig:example}
\end{figure}

State-of-the-art multimodal methods typically adopt Transformer-based backbones for joint feature modeling~\cite{zhang2023cmx,zhang2023delivering}. These models leverage cross-attention mechanisms for modality interaction and often incorporate domain adaptation to transfer semantics from labeled image data to sparsely labeled event streams~\cite{sun2022ess,xie2024cross}. While effective, such approaches are computationally demanding due to the quadratic complexity of self-attention and the high dimensionality of combined inputs. Moreover, most existing fusion strategies emphasize spatial-level alignment while underutilizing the unique temporal dynamics of event data, which can result in suboptimal cross-modal synergy and semantic inconsistency.

To overcome these challenges, we draw inspiration from Mamba~\cite{gu2023mamba}, a recently proposed state space model that supports input-conditioned, long-range sequence modeling with linear computational complexity. Mamba has demonstrated strong performance and scalability in a variety of vision tasks, including classification~\cite{liu2024vmamba,ma2024tinyvim}, detection~\cite{wang2025mamba}, and segmentation~\cite{ruan2024vm}. Motivated by these advantages, we present MambaSeg, a novel dual-branch semantic segmentation framework that leverages parallel Mamba encoders to model RGB images and event streams independently, enabling efficient and scalable multimodal representation learning. 

To reduce cross-modal ambiguity, we introduce the Dual-Dimensional Interaction Module (DDIM) to enhance the information exchange between different modalities, which fuses image and event features along both spatial and temporal dimensions. DDIM is composed of two key components: the Cross-Spatial Interaction Module (CSIM), which aligns spatial semantics by integrating dense texture features from images with structural edge cues from events; and the Cross-Temporal Interaction Module (CTIM), which exploits Mamba’s global modeling capacity to refine temporal dependencies via attention-guided fusion. This design ensures fine-grained, modality-aware feature integration while minimizing computational overhead.
We validate MambaSeg on two public benchmarks, DDD17 and DSEC. Our method achieves 77.56\% mIoU on DDD17 and 75.11\% mIoU on DSEC, setting new SOTA performance while maintaining high efficiency. These results highlight the strength of our MambaSeg for multimodal segmentation.

Our main contributions are summarized as follows:
\begin{itemize}
    \item We propose MambaSeg, a dual-branch semantic segmentation framework based on parallel Mamba encoders that model image and event modalities efficiently with long-range dependency modeling and linear complexity. 
    \item We design the DDIM module, which includes CSIM and CTIM for structured spatial-temporal fusion. This module jointly leverages Mamba and attention to enhance cross-modal feature alignment and reduce ambiguity. 
    \item We conduct extensive experiments on the DDD17 and DSEC datasets, where MambaSeg achieves SOTA performance while significantly reducing computational cost compared to Transformer-based baselines. 
\end{itemize}

\section{Related Work}

\subsection{Semantic Segmentation}
RGB-based semantic segmentation has advanced significantly with CNNs and Transformers. Encoder-decoder architectures like SegFormer~\cite{xie2021segformer} and SegNeXt~\cite{guo2022segnext} capture both spatial details and high-level semantics, achieving SOTA results. However, their dependence on conventional sensors limits robustness in fast motion, low-light, and high dynamic range scenarios, where motion blur and latency degrade performance.

To address these issues, event cameras have attracted interest for their high temporal resolution, low latency, and wide dynamic range. EV-SegNet~\cite{alonso2019ev} pioneered event-based segmentation. Due to scarce labeled event data, methods like ESS~\cite{sun2022ess} and CMESS~\cite{xie2024cross} employ unsupervised domain adaptation to transfer semantic knowledge from images. Recent Transformer-based models such as EvSegFormer~\cite{jia2023event} integrate motion priors into attention to exploit event dynamics. Spiking neural networks (SNNs), including Spike-BRGNet~\cite{long2024spike} and SLTNet~\cite{long_SLTNet}, have also been used to model temporal information efficiently.

RGB-event fusion leverages complementary spatial and temporal cues for robust perception. Hybrid frameworks like HALSIE~\cite{das2024halsie} and SpikingEDN~\cite{zhang2024accurate} combine ANNs and SNNs to capture static textures and dynamic edges. RGB-X methods such as CMX~\cite{zhang2023cmx} and CMNeXt~\cite{zhang2023delivering} employ multi-scale alignment and cross-attention for spatial fusion. EISNet~\cite{xie2024eisnet} further introduces gated attention and progressive recalibration for adaptive feature alignment.

However, many methods rely on computationally expensive Transformer attention and suffer from cross-modal ambiguity. To address this, we propose a cross-modal interaction module that enhances spatial-temporal complementarity, improving fusion quality and segmentation performance.

\begin{figure*}[t]
\centering
\includegraphics[width=0.8\textwidth]{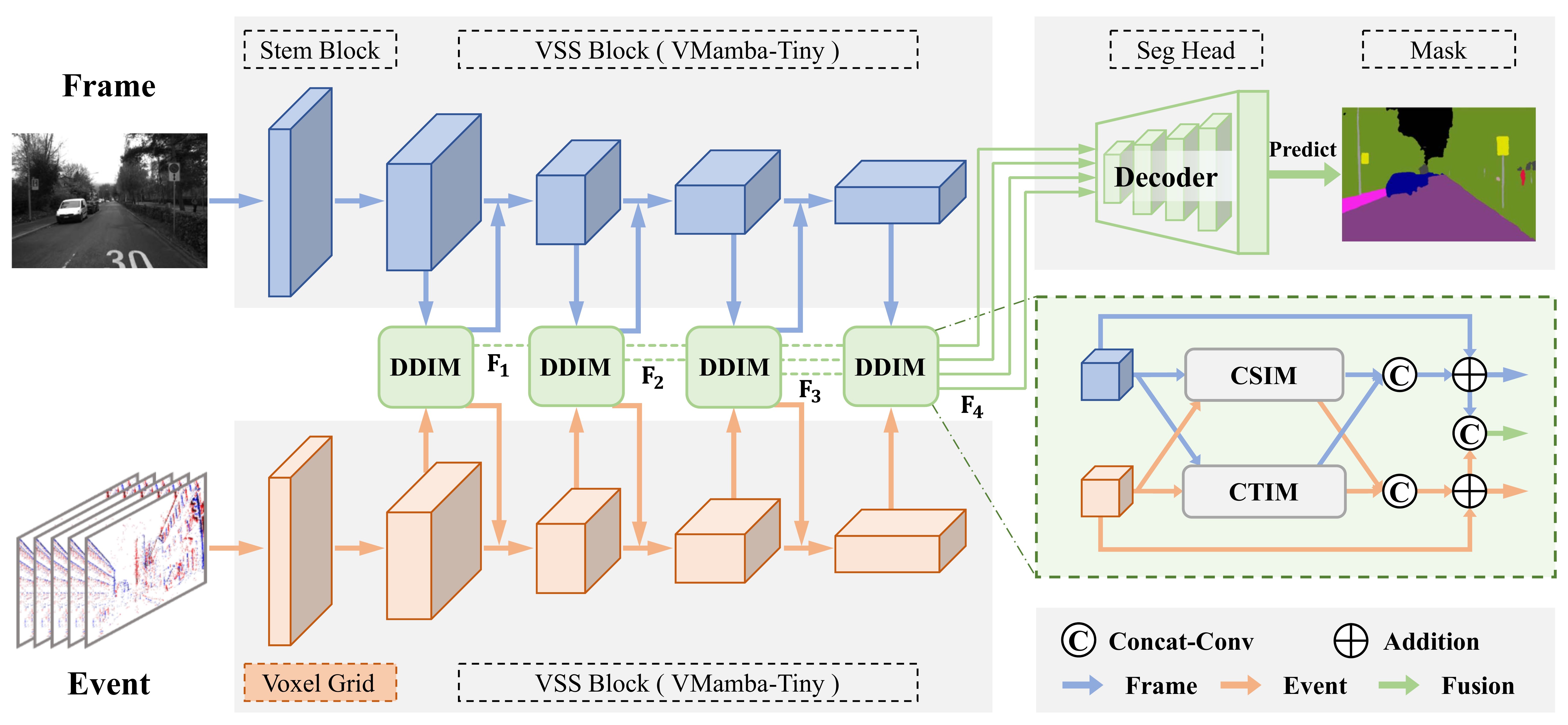}
\caption{Overview of the MambaSeg framework. MambaSeg consists of a dual-branch Mamba encoder and a Dual-Dimensional Interaction Module (DDIM), which includes CSIM and CTIM. Voxelized event and image streams are independently processed by multi-scale VSS Blocks. DDIM performs spatial-temporal fusion at each scale, and the fused features are iteratively fed into the encoders to enhance cross-modal consistency, followed by a decoder for semantic segmentation.
}
\label{fig:framework}
\end{figure*}

\subsection{State Space Models}
State Space Models (SSMs), introduced in S4~\cite{gu2021efficiently}, exhibit strong global modeling and long-range dependency handling, surpassing CNNs and Transformers in sequence tasks. Subsequent work~\cite{smith2022simplified} enhanced efficiency via parallelizable linear-complexity scans.
Mamba~\cite{gu2023mamba} advances this with an input-dependent selective scanning mechanism for dynamic sequence processing. Vision Mamba \cite{liu2024vmamba} adapts it for visual tasks through multi-directional state propagation, achieving strong image classification results. The framework has since succeeded in medical segmentation \cite{guo2024mambair} and point cloud analysis \cite{wang2025mamba}, demonstrating its versatility across vision applications.

However, the potential of Mamba in multimodal fusion, particularly in integrating dense RGB images with sparse event data, remains largely untapped. To address this, we propose a novel Mamba-Attention architecture designed for cross-modal segmentation, combining Mamba's efficient sequence modeling with structured spatial-temporal fusion for improved alignment and representation across modalities.

\section{Proposed Method: MambaSeg}

\subsection{Overview of the MambaSeg}
The proposed MambaSeg framework, depicted in Fig.~\ref{fig:framework}, consists of the dual-branch Mamba encoder and the DDIM. The DDIM comprises two integral components: the CSIM and the CTIM, designed to enhance multimodal feature fusion across spatial and temporal dimensions.

MambaSeg takes as input both image frames and asynchronous events. The image frames are represented as \( I \in \mathbb{R}^{C \times H \times W} \) as input. Meanwhile, the raw asynchronous event stream is transformed into a structured voxel grid to preserve spatio-temporal information. Each event is represented as a tuple \( e_i = (x_i, y_i, t_i, p_i) \), where \( (x_i, y_i) \) denotes the spatial location, \( t_i \) the timestamp, and \( p_i \in \{-1, +1\} \) the polarity. Given a fixed time window, we divide the time span into \( T \) discrete bins and accumulate the events into a voxel grid \( E \in \mathbb{R}^{T \times H \times W} \).
The voxel intensity at each spatio-temporal coordinate is computed as:
\begin{align}
E(t, x, y) &= \sum_{j=1}^{N} \delta(x_j = x, y_j = y) \cdot \delta(t_j \in B_t) \cdot p_j,
\end{align}
where \( B_t \) denotes the time interval of the \( t \)-th temporal bin, and \( \delta(\cdot) \) is the Kronecker Delta function. The resulting voxel grid encodes the spatial and temporal distribution of events and serves as the input to the event branch of our model.


The DDIM module is central to our framework. We use four-scale Visual State Space (VSS) Blocks \cite{liu2024vmamba} to encode image and event modalities independently. At each scale, DDIM facilitates cross-modal interaction and fusion. Its CSIM component leverages image spatial semantics and event edge cues for fine-grained spatial fusion, while CTIM operates along the temporal axis, aligning dynamic events with static visuals to enhance temporal modeling. After each stage, fused features are fed to subsequent encoders, progressively improving cross-modal consistency. Finally, the fused features across all scales are passed through a decoder to generate the final segmentation predictions.

\subsection{Cross Spatial Interaction Module}

\begin{figure}[t]
\centering
\includegraphics[width=0.45\textwidth]{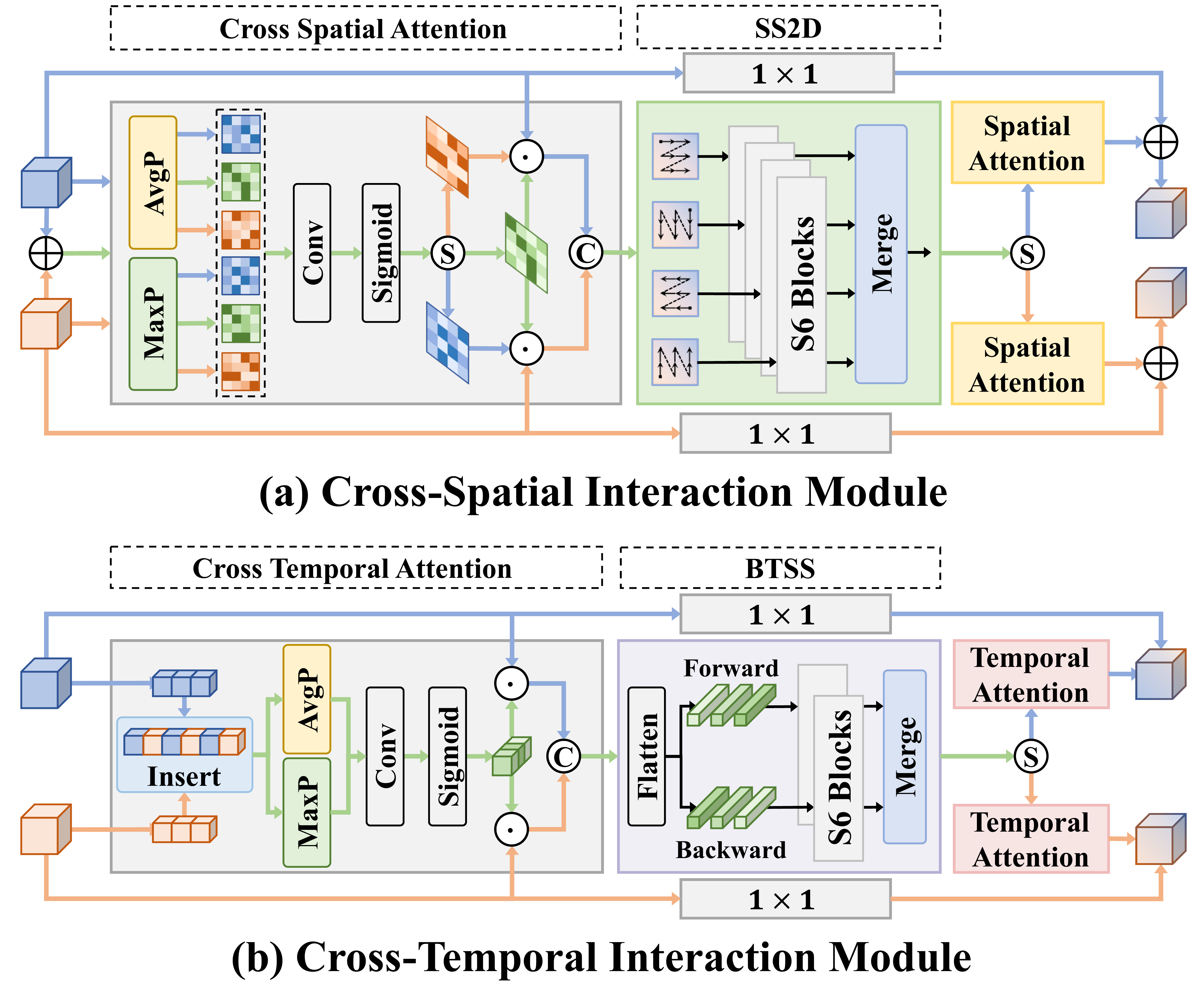}
\caption{The detail of (a) CSIM and (b) CTIM. }
\label{fig:csim}
\end{figure}

To fully exploit the edge cues from event data and the texture-rich information from images while reducing cross-modal ambiguity, we introduce the CSIM, as illustrated in Fig.~\ref{fig:csim} (a), which achieves efficient cross-modal interaction and fusion through three key components: cross-modal spatial attention, spatial refinement via SS2D, and modality-aware residual updates. Leveraging this design, it effectively integrates complementary features.

\paragraph{Cross-Modal Spatial Attention.}
To ensure dimensional compatibility for cross-modal fusion, we align the number of image channels with the number of event time steps at each feature extraction stage. Consequently, the event and image features at stage \( i \) are represented as \( E_i \in \mathbb{R}^{T \times H \times W} \) and \( I_i \in \mathbb{R}^{T \times H \times W} \), respectively. We first compute a shallow fusion feature via element-wise addition \( F^S_i = E_i + I_i \). To capture complementary spatial cues across modalities, we apply both average and max pooling on each of the features \( E_i \), \( I_i \), and \( F_i \), yielding a total of six spatial feature maps:
\begin{align}
X_i &= \big[ 
    \mathrm{AvgPool}(E_i),\ \mathrm{MaxPool}(E_i), \notag \\
  &\quad\ \mathrm{AvgPool}(I_i),\ \mathrm{MaxPool}(I_i), \notag \\
  &\quad\ \mathrm{AvgPool}(F_i),\ \mathrm{MaxPool}(F_i) 
\big] \in \mathbb{R}^{6 \times H \times W}.
\end{align}

These spatial maps are concatenated along the channel dimension and passed through two convolutional layers and a sigmoid activation to generate the spatial attention weight:
\begin{equation}
W^S = \sigma(\mathrm{Conv}_2(\mathrm{ReLU}(\mathrm{Conv}_1(X_i)))) \in \mathbb{R}^{3 \times H \times W},
\end{equation}
where \( \sigma \) is the sigmoid function. We divide \( W^S \) into three spatial attention maps, denoted as \( W^S_E \), \( W^S_I \), and \( W^S_F \), corresponding to \( E_i \), \( I_i \), and \( F_i \), respectively. Cross-modal interaction is then achieved by applying these attention maps. Specifically, event-guided attention sharpens geometric edges while image-guided attention enriches texture, facilitating deep interaction between the modalities: 
\begin{align}
E^S_c &= E_i \odot W^S_I \odot W^S_F, \notag \\
I^S_c &= I_i \odot W^S_E \odot W_F, \notag \\
F^S_c &= \mathrm{Concat}(E^S_c, I^S_c),
\end{align}
where \( \odot \) denotes element-wise multiplication and \( \mathrm{Concat} \) represents channel-wise concatenation.

\paragraph{Spatial Refinement via SS2D.}
To further enhance the spatially fused features, we feed \( F^S_c \) into the SS2D module:
\begin{equation}
F^S_s = \mathrm{SS2D}(F^S_c).
\end{equation}

Within SS2D, the feature map is unfolded into four directional sequences based on predefined patch configurations. Each sequence is independently processed by a dedicated S6 Block to capture long-range dependencies along its respective orientation. Resulting outputs are subsequently fused and reshaped into a 2D spatial feature map. This directional sequence modeling enables SS2D to capture diverse contextual cues across multiple spatial orientations effectively.

\paragraph{Modality-Aware Residual Update.}
The fused feature \( F^S_s \) is split back into separate image and event modality features: \( \{E^S_s, I^S_s\} = \mathrm{Split}(F^S_s) \). We then apply a spatial attention module \( \mathrm{SA}(\cdot) \) to each branch, followed by residual connections. This design allows the updated features to retain modality-specific characteristics while integrating contextually enriched information:
\begin{align}
E^S_{i+1} &= E_i + E^S_s \odot \mathrm{SA}(E^S_s), \notag \\
I^S_{i+1} &= I_i + I^S_s \odot \mathrm{SA}(I^S_s), 
\end{align}
where \( E^S_{i+1} \) and \( I^S_{i+1} \) denote the refined event and image features, which are propagated to the next stage of the network.

\subsection{Cross Temporal Interaction Module}

To exploit the inherent temporal dynamics of event data, we introduce CTIM, as shown in Fig.~\ref{fig:csim}(b). This module enables temporal cross-modality alignment through three components: cross-modal temporal attention, bi-directional temporal selective scanning, and modality-aware residual update. By aligning event features with corresponding image features, CTIM mitigates temporal inconsistencies and enhances complementarity between modalities.

\paragraph{Cross-Modal Temporal Attention.}
Given the event and image features at stage \( i \), denoted as \( E_i \in \mathbb{R}^{T \times H \times W} \) and \( I_i \in \mathbb{R}^{T \times H \times W} \), we construct a temporal fusion feature \( F^T_i \in \mathbb{R}^{2T \times H \times W} \) by temporally interleaving event and image features: 
\begin{equation}
F^T_i = \mathrm{Insert}(E_i, I_i),
\end{equation}
where event features are inserted between adjacent image channels along the temporal dimension.  This design preserves the temporal sequence while explicitly modeling interleaved dependencies between modalities. 

We then apply global max pooling and average pooling to the temporally interleaved feature $F^T_i$ to extract temporal response descriptors:
\begin{align}
F^T_{\mathrm{max}} &= \mathrm{MaxPool}(F^T_i),  &F^T_{\mathrm{avg}} &= \mathrm{AvgPool}(F^T_i).
\end{align}

These pooled features are projected into temporal attention weights through two consecutive \(1 \times 1\) convolution layers followed by a sigmoid activation:
\begin{align}
W^T_F = \sigma\Big( 
\mathrm{Conv}(F^T_{\mathrm{max}}) + \mathrm{Conv}(F^T_{\mathrm{avg}}) 
\Big) \in \mathbb{R}^{T \times 1 \times 1}.
\end{align}

The resulting attention weights \( W^T_F \) are broadcast and applied to the original modality inputs, producing temporally modulated representations:
\begin{align}
E^T_c &= E_i \odot W^T_F,&I^T_c &= I_i \odot W^T_F.
\end{align}

This attention mechanism emphasizes motion-sensitive features in the event stream and suppresses redundant or ambiguous static cues in the image stream, thereby promoting robust and discriminative temporal fusion across modalities.

\paragraph{Bi-Directional Temporal Selective Scan.}
To further model long-range dependencies, we concatenate the attended features as:
\begin{align}
F^T_c = \mathrm{Concat}(E^T_c, I^T_c)\in \mathbb{R}^{2T \times H\times W}, 
\end{align}
and then flatten the spatial to form a temporal sequence \( F^T_{\mathrm{flat}} \in \mathbb{R}^{2T \times HW} \). This sequence is processed in both forward and reverse directions using separate S6 blocks:
\begin{align}
F^T_{\mathrm{fwd}} &= \mathrm{S6}(F^T_{\mathrm{flat}}), &
F^T_{\mathrm{bwd}} &= \mathrm{S6}(\mathrm{Reverse}(F^T_{\mathrm{flat}})).
\end{align}

The outputs are summed and reshaped back to a 3D form to produce the temporally enriched feature map:
\begin{equation}
F^T_b = \mathrm{Reshape}(F^T_{\mathrm{fwd}} + F^T_{\mathrm{bwd}}).
\end{equation}

This bidirectional design enables the aggregation of temporal context from both past and future frames, resulting in more coherent and contextually aligned representations.

\paragraph{Modality-Aware Residual Update.}
Similar to CSIM, we split the fused features back into modality-specific branches, i.e., \( \{E^T_b, I^T_b\} = \mathrm{Split}(F^T_b) \). Each branch is refined using a temporal attention module \( \mathrm{TA}(\cdot) \), followed by a residual connection with the original input:
\begin{align}
E^T_{i+1} &= E_i + E^T_b \odot \mathrm{TA}(E^T_b), \notag \\
I^T_{i+1} &= I_i + I^T_b \odot \mathrm{TA}(I^T_b),
\end{align}
where \( E^T_{i+1}, I^T_{i+1} \) are the updated event and image features passed to the next stage. This update preserves the semantic identity of each modality while integrating complementary temporal cues, contributing to a progressively aligned and semantically enriched cross-modal representation.

\begin{table*}[t]
\begin{small}
\centering
\resizebox{0.95\textwidth}{!}
{
\begin{tabular}{lccccccccc}
\toprule
\multirow{2.5}{*}{\textbf{Method}} & \multirow{2.5}{*}{\textbf{Publication}} & \multirow{2.5}{*}{\textbf{Modality}} & \multirow{2.5}{*}{\textbf{Backbone}} & \multirow{2.5}{*}{\textbf{Representation}} & \multicolumn{2}{c}{\textbf{DDD17}} & \multicolumn{2}{c}{\textbf{DSEC}} \\
\cmidrule(lr){6-7} \cmidrule(lr){8-9}
 & & & & & \textbf{mIoU (\%)} & \textbf{Acc. (\%)} & \textbf{mIoU (\%)} & \textbf{Acc. (\%)} \\
\midrule
SegNeXt & NeurIPS'21 & Image & CNN & - & 71.46 & 95.97 & 71.55 & 94.89 \\
SegFormer & NeurIPS'22 & Image & Transformer & - & 71.05 & 95.73 & 71.99 & 94.97 \\
\midrule
EV-SegNet & CVPR'19 & Event & CNN & 6-Channel & 54.81 & 89.76 & 51.76 & 88.61 \\
ESS & ECCV'22 & Event & CNN & Voxel Grid & 61.37 & 91.08 & 51.57 & 89.25 \\
\midrule
ESS         & ECCV'22 & Image-Event  & CNN      & Voxel Grid   & 60.43  & 90.37 & 53.29 & 89.37 \\
EDCNet  & TITS'22 & Image-Event  & CNN & Voxel Grid   & 61.99  & 93.80 & 56.75 & 92.39 \\
HALSIE  & WACV'24 & Image-Event  & CNN+SNN & Voxel Grid   & 60.66  & 92.50 & 52.43 & 89.01 \\
Hybrid-Seg  & AAAI'25 & Image-Event  & CNN+SNN & Voxel Grid   & 67.31  & 95.07 & 66.57 & 94.27 \\
CMX         & TITS'23 & Image-Event  & Transformer     & Voxel Grid   & 71.88  & 95.64 & 72.42 & 95.07 \\
CMNeXt      & CVPR'23 & Image-Event  & Transformer     & Voxel Grid   & 72.67  & 95.74 & 72.54 & 95.10 \\
SE-Adapter & ICRA'24 & Image-Event & Transformer & MSP & 69.06 & 95.32 & 69.77 & 93.58 \\
EISNet      & TMM'24 & Image-Event  & Transformer     & AET          & 75.03  & 96.04 & 73.07 & 95.12 \\
\midrule
\textbf{MambaSeg} & Ours & Image-Event & Mamba & Voxel Grid & \textbf{77.56} & \textbf{96.33} & \textbf{75.10} & \textbf{95.71} \\
\bottomrule
\end{tabular}
}
\caption{Comparison with state-of-the-art semantic segmentation methods on DDD17 and DSEC datasets.}
\label{tab:Table1}
\end{small}
\end{table*}

\section{Experiments and Results}
\subsection{Datasets and Evaluation Metrics}
Experiments are conducted on two widely used event-based semantic segmentation datasets: DDD17 and DSEC-Semantic (DSEC). 
The DDD17 dataset~\cite{binas2017ddd17, alonso2019ev} contains paired event data and grayscale images captured from driving scenes using DAVIS sensors at a resolution of 346×260. Semantic annotations are generated using a pretrained segmentation model on synchronized images, covering six categories. The dataset comprises 15,950 training pairs and 3,890 testing pairs.

The DSEC dataset~\cite{sun2022ess} includes driving sequences with event streams and high-resolution RGB images (440×640), annotated with 11 fine-grained semantic categories. We follow the official data split, which consists of 8,082 training frames across eight sequences and 2,809 testing frames across three sequences, as well as the original preprocessing pipeline.

We evaluate segmentation performance using mean Intersection over Union (mIoU) and pixel accuracy. Model complexity is reported in terms of the number of trainable parameters and multiply-accumulate operations (MACs).

\subsection{Implementation Details}

\begin{figure*}[t]
\centering
\includegraphics[width=0.9\textwidth]{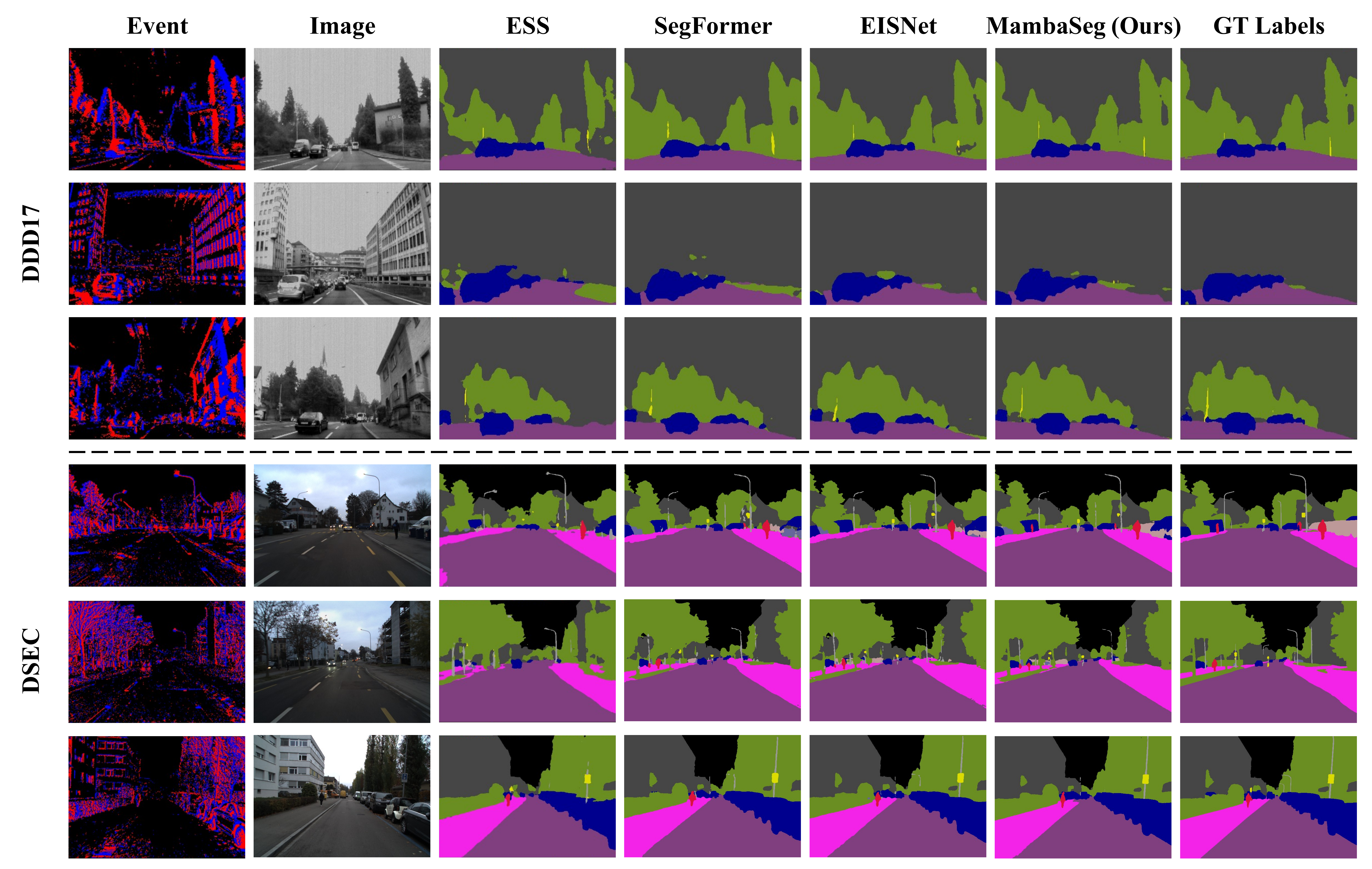}
\caption{
    Qualitative comparison of different segmentation methods on DDD17 and DSEC datasets.
}
\label{fig:vis}
\end{figure*}
We employ the VMamba-T model \cite{liu2024vmamba}, pretrained on ImageNet-1K, as the encoder for both the event and image branches. Each branch utilizes a four-stage encoder.
For the segmentation head, we adopt the MLP decoder architecture from SegFormer \cite{xie2021segformer}.

All models are implemented in PyTorch and trained on a single NVIDIA RTX-4090D GPU using the AdamW optimizer and cross-entropy loss. Training is conducted for 60 epochs on both datasets. On DDD17, we use an initial learning rate (\textit{lr}) of 2e-4 with a batch size (\textit{bs}) of 12; on DSEC, the \textit{lr} is set to 6e-5, with a \textit{bs} of 4.
Following prior work \cite{xie2024eisnet}, we construct voxel grids by segmenting the event stream into 10 intervals. For DDD17, segmentation is based on fixed 50 ms intervals, while for DSEC, each segment contains 100,000 events. To ensure fair comparison, we apply standard data augmentations, including random cropping, horizontal flipping, and random resizing.

\subsection{Results and Analysis}

We benchmark MambaSeg against SOTA segmentation methods, categorizing by input modality into: image-only, event-only, and event-image fusion. Image-only methods include SegFormer~\cite{xie2021segformer} and SegNeXt~\cite{guo2022segnext}. Event-only methods comprise EV-SegNet~\cite{alonso2019ev} and ESS~\cite{sun2022ess}. Event-image fusion methods include EDCNet~\cite{zhang2021exploring}, HALSIE~\cite{das2024halsie},Hybrid-Seg~\cite{li2025efficient}, CMX~\cite{zhang2023cmx}, CMNeXt~\cite{zhang2023delivering}, SE-Adapter~\cite{yao2024sam}, and EISNet~\cite{xie2024eisnet}.


\begin{table}[t]
\begin{small}
\centering
\setlength{\tabcolsep}{1pt} 
 \resizebox{.45\textwidth}{!}{
\begin{tabular}{lcccc}
\toprule
\textbf{Method} & \textbf{Backbone} & \parbox{0.5in}{\centering \textbf{Params\\(M)}} & \parbox{0.5in}{\centering \textbf{MACs\\(G)}} & \parbox{0.5in}{\centering \textbf{mIoU\\(\%)}} \\
\midrule
Ev-SegNet & CNN         & 29.09 & 73.62  & 54.81 \\
EDCNet & CNN         & \textbf{23.06} & \textbf{6.14}  & 61.99 \\
SegFormer  & Transformer & 51.54          & 19.78          & 73.25 \\
SwiftFromer& Transformer & 64.61          & 23.88          & 74.31 \\
CMX        & Transformer & 66.56          & 16.29          & 71.88 \\
CMNeXt     & Transformer & 58.68          & 16.32          & 72.67 \\
EISNet     & Transformer & 34.39          & 17.30       & \underline{75.03} \\
\midrule
\textbf{MambaSeg (Ours)} & Mamba    & \underline{25.44} & \underline{15.59} & \textbf{77.56} \\
\bottomrule
\end{tabular}
}
\caption{Model complexity on DDD17 dataset.}
\label{tab:compute}
\end{small}
\end{table}

\subsubsection{Quantitative Evaluation.}
Table~\ref{tab:Table1} illustrates that MambaSeg outperforms SOTA methods on the DDD17 and DSEC benchmarks. On the DDD17 dataset, MambaSeg achieves a mIoU of 77.56\% and an accuracy of 96.33\%, surpassing the previous best method, EISNet, by 2.53\% in mIoU. Similarly, on the DSEC dataset, MambaSeg achieves the highest mIoU of 75.10\% and accuracy of 95.71\%, exceeding EISNet by 2.03\% in mIoU. These results underscore MambaSeg’s superior performance in event-image fusion semantic segmentation. The enhanced performance is attributed to three key innovations: (1) Parallel Mamba encoders with global receptive fields effectively capture rich feature representations from both modalities. (2) The CSIM facilitates fine-grained spatial fusion by leveraging complementary image textures and event edges, improving robustness and spatial consistency. (3) The CTIM enhances temporal coherence in event streams through cross-modal temporal fusion, mitigating cross-modal ambiguity.

Beyond segmentation accuracy, we also evaluate computational efficiency on the DDD17 dataset (Table~\ref{tab:compute}). Compared to Transformer-based fusion methods such as CMX and EISNet, MambaSeg achieves the best mIoU (77.56\%) with significantly fewer parameters (25.44M) and moderate MACs (15.59G). Relative to CNN-based methods, MambaSeg delivers substantially higher accuracy with comparable or lower computational cost. These results highlight the efficiency of the Mamba architecture and demonstrate that MambaSeg achieves an excellent balance between performance and efficiency.

\subsubsection{Qualitative Evaluation.}

As illustrated in Fig.~\ref{fig:vis}, we present qualitative segmentation results on the DDD17 and DSEC datasets, comparing MambaSeg with ESS (event-only), SegFormer (image-only), and the SOTA fusion method EISNet. Due to the inherent sparsity of event data, event-only methods often fail to recover complete semantic regions. In contrast, image-only methods such as SegFormer struggle with small object segmentation (e.g., pedestrians) under challenging lighting or cluttered backgrounds, as they are less sensitive to dynamic changes.

Compared to EISNet, MambaSeg produces more accurate segmentation of small and complex objects, such as pedestrians and traffic signs. This highlights the effectiveness of our dual-dimensional fusion strategy, which combines the temporal dynamics of event data with the rich texture information of images along both spatial and temporal axes.

\subsection{Ablation Study}

To assess the contribution of each component, we conduct ablation studies on the DDD17 dataset, focusing on the proposed CSIM, CTIM, and their respective sub-components.

\subsubsection{Comparison of Cross-Modal Fusion Methods.} We assess the effectiveness of DDIM by comparing it with representative fusion strategies under a unified setup. All methods adopt the same VMamba-T encoder and are applied after feature encoding, with the rest of the architecture unchanged. We use element-wise addition as the baseline and compare with FFM~\cite{zhang2023cmx}, MRFM~\cite{xie2024eisnet}, CSF~\cite{li2025efficient}, and our DDIM.
As shown in Table~\ref{tab:fusion}, DDIM outperforms all competitors, achieving 77.56\% mIoU and 96.33\% pixel accuracy on DDD17. This improvement stems from DDIM’s dual-axis fusion, which effectively aligns spatial and temporal features to enhance cross-modal complementarity. In contrast, element-wise addition lacks any interaction modeling (74.38\% mIoU), while FFM, MRFM, and CSF offer limited fusion capabilities, lagging behind DDIM by 1.12\%, 1.37\%, and 0.91\% mIoU, respectively. By aligning modalities along both spatial and temporal dimensions, DDIM enables more effective feature integration and leads to superior segmentation performance.

\begin{table}[t]
\begin{small}
\centering
\begin{tabular}{lcc}
\toprule
\textbf{Fusion Method} & \textbf{mIoU (\%)} & \textbf{Acc. (\%)} \\
\midrule
Baseline & 74.38 & 95.96 \\
FFM~\cite{zhang2023cmx}             & 76.44 & 96.06 \\
MRFM~\cite{xie2024eisnet}             & 76.19 & 95.97 \\
CSF~\cite{li2025efficient}              & 76.65 & 96.22 \\
\textbf{DDIM (Ours)}        & \textbf{77.56} & \textbf{96.33} \\
\bottomrule
\end{tabular}
\caption{Comparison of different cross-modal fusion methods on DDD17 dataset.}
\label{tab:fusion}
\end{small}
\end{table}

\subsubsection{Effect of CSIM and CTIM.}
We first evaluate the individual and combined impact of CSIM and CTIM. As shown in Table~\ref{tab:ablation_ddim}, removing either module results in a performance drop, while combining both yields the highest mIoU of 77.56\% and accuracy of 96.33\%. These results underscore the complementary strengths of spatial and temporal fusion in improving segmentation performance.

\begin{table}[!t]
\begin{small}
\centering
\begin{tabular}{ccccc}
\toprule
\textbf{CSIM} & \textbf{CTIM} & \textbf{mIoU (\%)} & \textbf{Acc. (\%)} \\
\midrule
\ding{55} & \ding{55} & 74.38 & 95.96 \\
\ding{55} & \ding{51} & 76.20 & 96.06 \\
\ding{51} & \ding{55} & 76.32 & 96.25 \\
\ding{51} & \ding{51} & \textbf{77.56} & \textbf{96.33} \\
\bottomrule
\end{tabular}
\caption{Effectiveness of CSIM and CTIM on DDD17. Removing either CSIM or CTIM leads to a noticeable drop in performance, confirming their complementary roles in enhancing segmentation accuracy.}
\label{tab:ablation_ddim}
\end{small}
\end{table}

\begin{table}[!t]
\begin{small}
\centering
\setlength{\tabcolsep}{2.5pt}
\begin{tabular}{lcccccc}
\toprule
\textbf{Variation}     & \textbf{CSA} & \textbf{SS2D} & \textbf{SA} & \textbf{mIoU (\%)} & \textbf{Acc. (\%)} \\
\midrule
CSA + SS2D            & \ding{51} & \ding{51} & \ding{55} & 76.47 & 96.24 \\
SS2D + SA             & \ding{55} & \ding{51} & \ding{51} & 76.59 & 96.26 \\
CSA + SA              & \ding{51} & \ding{55} & \ding{51} & 76.71 & 96.24 \\
Full CSIM               & \ding{51} & \ding{51} & \ding{51} & \textbf{77.56} & \textbf{96.33} \\
\bottomrule
\end{tabular}
\caption{Component-wise ablation of CSIM. All components contribute positively to performance, with the complete CSIM achieving the highest accuracy and mIoU.}
\label{tab:csim-ablation}
\end{small}
\end{table}

\begin{table}[!t]
\centering
\begin{small}
\setlength{\tabcolsep}{2.5pt}
\begin{tabular}{lcccccc}
\toprule
\textbf{Variation}     & \textbf{CTA} & \textbf{BTSS} & \textbf{TA} & \textbf{mIoU (\%)} & \textbf{Acc. (\%)} \\
\midrule
CTA + BTSS            & \ding{51} & \ding{51} & \ding{55} & 76.49 & 96.15 \\
BTSS + TA             & \ding{55} & \ding{51} & \ding{51} & 76.66 & 96.17 \\
CTA + TA              & \ding{51} & \ding{55} & \ding{51} & 76.53 & 96.14 \\
Full CTIM               & \ding{51} & \ding{51} & \ding{51} & \textbf{77.56} & \textbf{96.33} \\
\bottomrule
\end{tabular}
\caption{Component-wise ablation of CTIM. All temporal fusion components contribute to performance gains, with the full CTIM achieving the best results.}
\label{tab:ctim-ablation}
\end{small}
\end{table}

\subsubsection{Ablation on CSIM Components.}
We further break down CSIM into three key components: Cross Spatial Attention (CSA), 2D Selective Scan (SS2D), and Spatial Attention (SA). As shown in Table~\ref{tab:csim-ablation}, each component contributes to the overall performance, with the full CSIM achieving the highest mIoU and accuracy. Notably, the combination of CSA and SA yields substantial improvements, highlighting their synergistic effect in capturing both fine-grained spatial structure and broader contextual semantics.

\subsubsection{Ablation on CTIM Components.}
We then conduct an ablation study on CTIM, which comprises Cross Temporal Attention (CTA), Bi-Directional Temporal Selective Scan (BTSS), and Temporal Attention (TA). As shown in Table~\ref{tab:ctim-ablation}, each component contributes meaningfully to performance. The complete CTIM configuration achieves the best results, demonstrating that jointly modeling temporal dynamics and alignment enhances the effectiveness of temporal fusion.

\section{Conclusion}
In this work, we proposed MambaSeg, a novel dual-branch framework for multimodal semantic segmentation, built upon the efficient and scalable Mamba architecture. To address the high computational demands of Transformer-based models and the difficulty of fusing RGB images with sparse event data, we introduced the DDIM, which incorporates both CSIM and CTIM components. DDIM enables fine-grained, complementary fusion by jointly aligning spatial details and temporal dynamics across modalities.
Extensive evaluations on the DDD17 and DSEC datasets show that MambaSeg achieves SOTA performance while significantly reducing model complexity, offering an excellent trade-off between segmentation accuracy and efficiency. 

Our future work aims to deploy MambaSeg on real-world robotic platforms to validate its practical effectiveness in resource-constrained, dynamic environments.

\section{Acknowledgments}


This work is jointly supported by the National Natural Science Foundation of China (No. 42474027, 62403085, 42174050, 62322601), China Postdoctoral Science Foundation (No. 2023M740402), 
and Fundamental Research Funds for the Central Universities (No. 2024IAIS-QN017, 2025CDJZDGF001).

\bibliography{aaai2026}

\end{document}